\documentclass[11pt]{article}

\usepackage[utf8]{inputenc}
\usepackage[T1]{fontenc}
\usepackage{times}
\usepackage{microtype}

\usepackage{makecell}
\usepackage{mathtools} 
\usepackage{amssymb}
\usepackage{amsthm}
\usepackage{graphicx}
\usepackage{subcaption}
\usepackage{booktabs}
\usepackage{multirow}
\usepackage{tabularx}
\usepackage{makecell}
\usepackage{dsfont}
\usepackage{bbm}

\usepackage{xcolor}
\usepackage{url}
\usepackage{hyperref}
\usepackage[capitalize,noabbrev]{cleveref}

\usepackage{microtype}
\usepackage{xspace}
\usepackage{adjustbox}

\usepackage{algorithm}
\usepackage{algpseudocode}
\usepackage{pifont}
\usepackage{graphicx}

\usepackage{enumitem}
\newcommand{\cmark}{\textcolor{green!60!black}{\ding{51}}}
\newcommand{\xmark}{\textcolor{red!70!black}{\ding{55}}}
\newcommand{\algcomment}[1]{\Statex \textcolor{gray!90}{\textit{#1}}}

\usepackage{xcolor}
\PassOptionsToPackage{table}{xcolor}
\usepackage{colortbl}
\definecolor{lightblue}{RGB}{220,235,255} 

\newtheorem{theorem}{Theorem}

\newtheorem{assumption}{Assumption}

\newtheorem{proposition}{Proposition}

\newcommand{\ours}{\texttt{FedWQ-CP}\xspace}

\theoremstyle{plain}

\theoremstyle{definition}

\theoremstyle{remark}

\title{Conformalized Neural Networks for Federated Uncertainty Quantification under Dual Heterogeneity}

\author{
\makebox[\textwidth][c]{%
Quang-Huy Nguyen$^{1}$ \quad
Jiaqi Wang$^{1,}$\textsuperscript{*} \quad
Wei-Shinn Ku$^{1,}$\textsuperscript{*}}\\[0.6em]
\makebox[\textwidth][c]{%
$^{1}$Department of Computer Science and Software Engineering, Auburn University}\\[0.4em]
\makebox[\textwidth][c]{%
\small\textsuperscript{*}Co-corresponding authors}\\[0.3em]
\makebox[\textwidth][c]{%
\footnotesize
\texttt{hqn0001@auburn.edu} \quad
\texttt{jqwang@auburn.edu} \quad
\texttt{wzk0004@auburn.edu}}
}

\date{}

\begin{document}
\maketitle

\begin{abstract}
Federated learning (FL) faces challenges in uncertainty quantification (UQ).
Without reliable UQ, FL systems risk deploying overconfident models at under-resourced agents, leading to silent local failures despite seemingly satisfactory global performance.
Existing federated UQ approaches often address data heterogeneity or model heterogeneity in isolation, overlooking their joint effect on coverage reliability across agents. Conformal prediction is a widely used distribution-free UQ framework, yet its applications in heterogeneous FL settings remains underexplored. 
We provide \ours, a simple yet effective approach that balances empirical coverage performance with efficiency at both global and agent levels under the dual heterogeneity. 
\ours performs agent–server calibration in a single communication round.
On each agent, conformity scores are computed on calibration data and a local quantile threshold is derived. Each agent then transmits only its quantile threshold and calibration sample size to the server. The server simply aggregates these thresholds through a weighted average to produce a global threshold.
Experimental results on seven public datasets for both classification and regression demonstrate that \ours empirically maintains agent-wise and global coverage while producing the smallest prediction sets or intervals.
\end{abstract}

\section{Introduction}


\begin{figure*}[t]
\centering
    \includegraphics[width=1\textwidth]{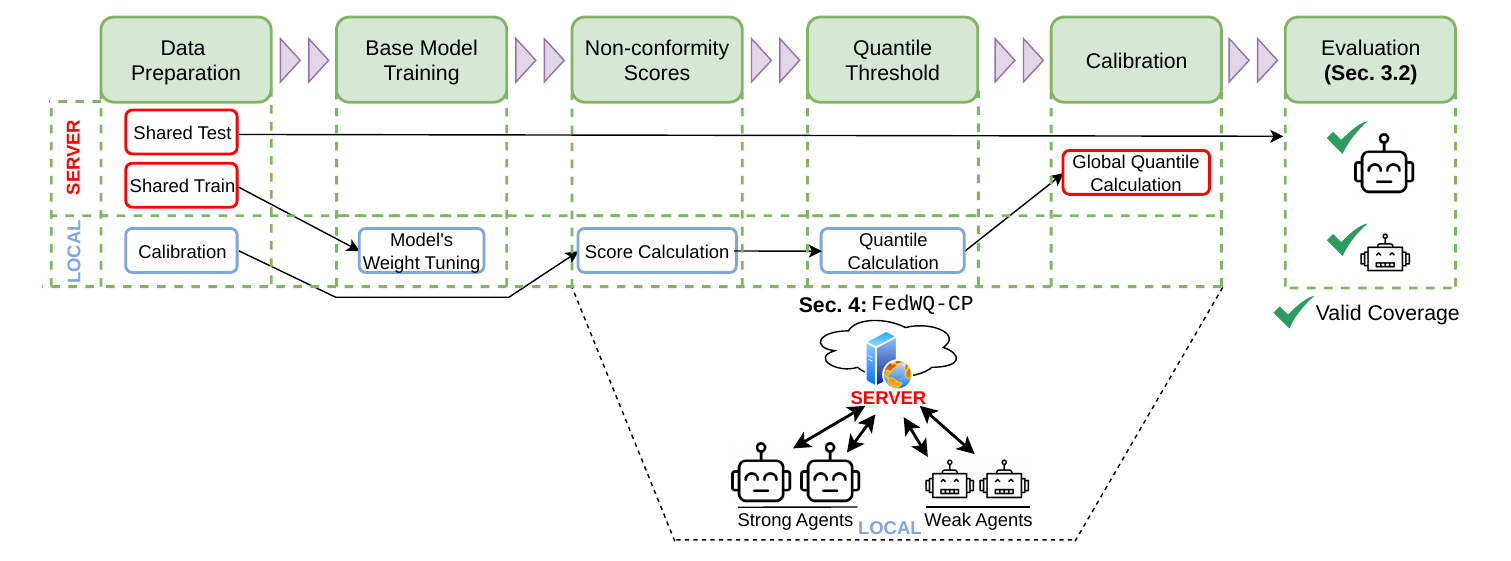}
    \caption{Framework of \ours. Strong agents denote higher-capacity architectures trained more extensively, while weak agents use lower-capacity architectures with reduced predictive strength. See Appendix~\ref{app:study_design} for details.}
    \label{fig:figure2}
\end{figure*}

In high-stakes federated learning (FL) systems, uncertainty quantification (UQ) informs critical decisions, yet data and model heterogeneity fundamentally reshape how uncertainty is generated and interpreted across agents. Consider a diagnostic system deployed across multiple hospitals. Each hospital trains its own predictor, potentially using different architectures and exhibiting varying predictive performance. At the same time, hospitals serve distinct patient populations, inducing distribution shifts and unequal dataset sizes.
Consequently, well-resourced hospitals with abundant data may achieve tight or even over-coverage, while smaller hospitals with limited data may experience systematic under-coverage$-$even when global coverage (the average of agent-wise marginal coverage across the federation) appears satisfactory\footnote{e.g., a 100\% coverage rate for a strong model (\underline{\textit{over-coverage}}) can offset a 90\% rate for a weak model (\underline{\textit{under-coverage}}), resulting in a 95\% average (\underline{\textit{exact coverage}}). Such averaging conceals silent failures to satisfy weak-agent-level coverage guarantees.}. 
As a result, the same empirical coverage level (e.g., 95\%) can correspond to substantially different reliability across sites (Figure~\ref{fig:fedavgqcp_ablation}). Such disparities can lead to inconsistent triage decisions, unequal escalation of care, and difficulty interpreting uncertainty across institutions.
Meanwhile, privacy and regulatory constraints prohibit the sharing of raw data or model parameters, while heterogeneous predictors further complicate direct standardization across agents.

FL~\cite{mcmahan2017communication}, a system widely adopted in the medical domain, addresses geographic health disparities by enabling collaborative model training without centralizing sensitive data~\cite{xu2022closing,wang2022towards,jiang2023fair,wang2023federated,wang2025asymmetrical}. Conformal prediction (CP)~\cite{vovk2005algorithmic} provides a distribution-free UQ framework with finite-sample marginal coverage guarantees. Given a miscoverage level $\alpha \in (0,1)$, CP constructs prediction sets or intervals with coverage at least $1-\alpha$ under exchangeability. 
However, CP applications in heterogeneous FL settings remain underexplored because classical guarantees assume exchangeability between calibration and test samples for a fixed predictor, an assumption that breaks down when agents operate on distinct data distributions and deploy heterogeneous models.

To resolve the challenges discussed above, this paper introduces a novel, powerful, yet straightforward \textbf{\underline{fed}}erated \textbf{\underline{w}}eighted \textbf{\underline{q}}uantile \textbf{\underline{c}}onformal \textbf{\underline{p}}rediction framework, \ours, a federated calibration method that aggregates locally computed conformal quantiles via weighted averaging as depicted in Figure~\ref{fig:figure2}. 

The \ours framework applies to diverse input modalities without imposing structural constraints on cross-agent data distributions or dataset sizes. Using a shared training data, each agent independently trains and freezes its base predictor, allowing for heterogeneous architectures and predictive strengths. The calibration data are then used to compute nonconformity scores and derive a local conformal quantile threshold. In a single communication round, each agent transmits only its local quantile and calibration sample size to the server. The server performs calibration-size-weighted aggregation to obtain a global threshold, as detailed in Section~\ref{sec:methods}. This global threshold is subsequently broadcasted to all agents for evaluation on a shared global test set, as outlined in Section~\ref{sec:evaluation}. This design targets empirical agent-level coverage while enabling efficient calibration under joint data and model heterogeneity. Our empirical investigation spans seven benchmark datasets across two tasks, classification and regression, and compares against state-of-the-art federated UQ baselines. The results consistently show that \ours empirically attains reliable global and agent-level coverage while substantially reducing inefficiency.


\section{Related Works} 
\label{sec:related_works}

\begin{table*}[t]
\centering
\caption{
Comparison of UQ methods in FL settings.
\textbf{Classification} and \textbf{Regression} indicate whether the method is directly applicable to classification and regression tasks, respectively.
\textbf{Federated} indicates whether calibration can be performed in a decentralized manner without sharing raw calibration data.
\textbf{One-shot} indicates whether calibration requires only a single round of communication between agents and the server.
\textbf{No Iter. Opt.} (No Iterative Optimization) indicates that the method does not rely on solving a federated optimization problem (e.g., gradient-based quantile estimation) during calibration.
\textbf{No Het. Assump.} (No Heterogeneity Assumption) indicates that the method does not require explicitly specifying or estimating a structural shift model (e.g., label shift or density ratios) across agents.
A \cmark indicates that the method satisfies the stated property.
FedCP-QQ computes the global calibration threshold via a quantile-of-quantiles scheme approximating the pooled mixture quantile, whereas \ours aggregates local conformal quantiles using calibration-size-weighted averaging.
}
\setlength{\tabcolsep}{4.2pt}
\renewcommand{\arraystretch}{1.03}
\resizebox{\textwidth}{!}{
\begin{tabular}{lcccccc}
\toprule
\textbf{Method} 
& \textbf{Classification} 
& \textbf{Regression} 
& \textbf{Federated} 
& \textbf{One-shot} 
& \textbf{No Iter. Opt.}
& \textbf{No Het. Assump.} \\
\midrule
SplitCP        & \cmark & \cmark & \xmark & \xmark & \cmark & \cmark \\
FCP            & \cmark & \cmark & \cmark & \xmark & \cmark & \cmark \\
CPhet          & \cmark & \cmark & \cmark & \xmark & \cmark & \xmark \\
DP-FedCP       & \cmark & \xmark & \cmark & \xmark & \xmark & \xmark \\
FedCP-QQ       & \cmark & \cmark & \cmark & \cmark & \cmark & \cmark \\
\midrule
\rowcolor{lightblue}
\textbf{\ours} & \cmark & \cmark & \cmark & \cmark & \cmark & \cmark \\
\bottomrule
\end{tabular}
}
\end{table*}

\textbf{Split Conformal Prediction.}
Split conformal prediction (SplitCP) \cite{oliveira2024split} is the canonical distribution-free framework for UQ under exchangeability. It constructs prediction sets using nonconformity scores computed on a calibration set and applies to both regression and classification via standard score functions such as conformalized quantile regression~\cite{romano2019conformalized} (CQR) or adaptive prediction set~\cite{romano2020classification} (APS). For fairness and comparability with prior work, we explicitly adapt SplitCP to the FL setting as a reference method, while recognizing that it was not originally designed for federated UQ.

\textbf{Federated Conformal Prediction.}
Federated Conformal Prediction (FCP) \cite{lu2023federated} extends CP to FL under partial exchangeability, aggregating calibration scores across agents to compute a shared global threshold. Its guarantees hold with respect to the global mixture distribution rather than individual agents. FedCP-QQ \cite{humbert2023oneshot} further improves communication efficiency by computing a quantile-of-quantiles in a one-shot manner. Both approaches enable decentralized calibration without structural distributional assumptions, but do not explicitly address heterogeneous score scaling across diverse models.

\textbf{Heterogeneity-Aware Conformal Methods.}
CPhet \cite{plassier2024efficient} and DP-FedCP~\cite{plassier2023labelshift} incorporate importance-weighted quantiles to handle statistical heterogeneity. CPhet corrects for covariate shift via density-ratio estimation and provides high-probability guarantees under modeling assumptions. DP-FedCP focuses on label shift and estimates weighted quantiles through a federated optimization procedure. These methods improve robustness under structured distributional shift but introduce additional estimation complexity and communication overhead.

\textbf{Positioning of \ours.}
\ours is designed to simultaneously satisfy all practical requirements of UQ in FL systems. It is directly applicable to both classification and regression tasks (\textbf{Classification} and \textbf{Regression}) without modification of the underlying conformal framework. The method is inherently federated, requiring no centralized access to raw calibration data (\textbf{Federated}), and achieves fully decentralized calibration in a single communication round (\textbf{One-shot}). Unlike optimization-based approaches, \ours does not rely on iterative federated gradient procedures or auxiliary training objectives for quantile estimation (\textbf{No Iter. Opt.}), thereby reducing computational and communication overhead. Importantly, \ours does not require explicit modeling or estimation of distributional shift (e.g., density ratios or label shift parameters) across agents for its implementation (\textbf{No Het. Assump.}). This combination of general applicability, communication efficiency, and the absence of explicit structural modeling assumptions makes \ours particularly well-suited for highly heterogeneous FL deployments, including heterogeneous agent models, unequal dataset sizes, and distributional imbalance.

\section{PRELIMINARIES}
\label{sec:prelims}

We consider a FL system with $M$ agents. 
We formalize the joint probability space over the augmented random variable $(K, X, Y)$,
where $K \in \{1, \dots, M\}$ denotes a random agent index,
drawn with probability $\mathbb{P}(K=k) = n_k / N$,
and conditional on $K=k$, $(X,Y) \sim P_k$.

Each agent $k$ deploys a local predictor $f_k$ and possesses a calibration set
$\mathcal{D}^{\text{cal}}_k = \{(x_{k,i}, y_{k,i})\}_{i=1}^{n_k}$.
We assume agent-wise exchangeability:

\begin{assumption}[Calibration-to-Test Shift Only] \label{assump:agent_exchange}
All agents train their predictors on a shared global training set, and evaluation is performed on a shared global test distribution $P_{\mathrm{test}}$.
For each agent $k$, the calibration samples are i.i.d.\ draws from a client-specific distribution $P_k^{\mathrm{cal}}$ induced by a Dirichlet partition of the global data.

This controlled design isolates calibration heterogeneity while 
keeping training and testing distributions shared across agents. 
It is intended to stress-test federated calibration under joint 
data and model heterogeneity, rather than to model fully general 
cross-silo training heterogeneity.

We allow $P_k^{\mathrm{cal}} \neq P_{\mathrm{test}}$, so calibration-to-test distribution shift may be present.
No additional shift is introduced in training or testing across agents.
\end{assumption}

\paragraph{Quantifying Dual Heterogeneity.} 
In our setting, heterogeneity is localized to the calibration and model training phases:
\begin{itemize}
    \item \textbf{Data Heterogeneity (Dirichlet Calibration Shift):} Each agent’s calibration set $\mathcal{D}^{cal}_k$ is sampled via a Dirichlet partition $\text{Dir}(\beta)$ applied to the global dataset. Thus, $q_k$ is computed on a label-imbalanced or covariate-skewed calibration subset, while coverage is evaluated on the shared global test distribution. Training and testing distributions remain common across agents; only calibration distributions differ.
    \item \textbf{Model Heterogeneity (Architecture \& Strength):} Agents deploy diverse architectures $\mathcal{F} = \{f_{arch_1}, f_{arch_2}, \dots, f_{arch_k}\}$ and varying training intensities $E_k$. 
    
    Agents may differ both in predictive strength and in the distribution of their induced nonconformity scores. These differences motivate performing calibration locally so that each model’s internal uncertainty scale is normalized through its own quantile threshold $q_k$.
\end{itemize}

A significant challenge in federated UQ is that diverse architectures (e.g., shallow CNNs vs. deep ResNets) produce softmax outputs with varying "temperatures" and scales, making their raw scores $S(X,Y)$ incomparable. By performing local calibration, each agent $k$ maps its internal model uncertainty into a quantile threshold $q_k$. 


Importantly, quantiles are invariant to monotone transformations of the scores within each agent.
Therefore, each local conformal quantile $q_k$ reflects a rank-based threshold relative to the score distribution induced by $f_k$ on $P_k$.
However, the numerical values of $q_k$ across agents are not generally comparable without further structural assumptions.
In this work, we treat weighted averaging of $\{q_k\}$ as a communication-efficient aggregation heuristic rather than as an exact surrogate for pooled conformal calibration.
As a result, it partially mitigates differences
in score scaling induced by heterogeneous model architectures.
However, $q_k$ remains expressed in the original score scale
of agent $k$, and therefore does not create a shared probability scale
across agents by itself.
By transmitting both the threshold $q_k$ and the calibration sample size $n_k$ to the server, \ours performs a weighted aggregation that accounts for both predictive strength (via the value of $q_k$) and statistical reliability (via the weight $n_k/N$). This allows the server to synthesize a global uncertainty boundary
that empirically stabilizes coverage across agents with heterogeneous architectures and predictive strengths.

\textbf{Remark} (Variance and Sample Size).
Weak predictors typically induce higher-variance nonconformity score distributions,
which increases the variance of their empirical quantile estimator.
Since the asymptotic variance of a sample quantile scales as
$\mathcal{O}(1/n_k)$, agents with small calibration sets produce
statistically noisier threshold estimates. By weighting the global threshold by $n_k$, \ours ensures that the global boundary $\hat{q}$ is not disproportionately skewed by agents with very small or statistically unrepresentative calibration sets, providing a robust "safety margin" against local model under-performance.

\subsection{Conformal Prediction}
CP produces a prediction set $\mathcal{C}(X)$ such that $P(Y \in \mathcal{C}(X)) \geq 1-\alpha$. Given calibration scores $V_1,\dots,V_n$, let 
$V_{(1)} \le \cdots \le V_{(n)}$ denote their order statistics.
Define the split conformal threshold $\widehat q = V_{(\lceil (n+1)(1-\alpha)\rceil)}$.
Equivalently, $\widehat q = \widehat F^{-1}(\tau)$ with 
$\tau = \lceil (n+1)(1-\alpha)\rceil / n$,
where $\widehat F$ is the empirical CDF.
In this work, we utilize APS for classification and CQR for regression to compute scores $V$. Unlike centralized CP, our setting requires a global threshold $\hat{q}$ that maintains validity across $M$ heterogeneous agents.

\subsection{Evaluation Metrics}
\label{sec:evaluation}
Let $\mathcal{D}^{\mathrm{test}}$ be the global test set. We evaluate methods based on:
\begin{enumerate}
    \item \textbf{Coverage:} Empirical rate $\mathrm{Cov} = \frac{1}{|\mathcal{D}^{\mathrm{test}}|} \sum \mathbb{I}\{ Y_i \in \mathcal{C}(X_i) \}$, assessed at both the agent-level ($\text{Cov}_k$) and global-level.
    \item \textbf{Efficiency:} The expected size of the prediction set:
    $\text{Eff}(\mathcal{C}) = \mathbb{E}_{X \sim P_{\text{test}}} [\text{size}(\mathcal{C}(X))]$, where $\text{size}(\cdot)$ is the cardinality for classification or interval length for regression. High-performing methods must minimize efficiency while satisfying $\text{Cov} \geq 1-\alpha$.
\end{enumerate}

In our simulated FL setting, the global test set corresponds to the original centralized test split, which is used solely for evaluation and is not accessible during calibration.
\section{METHODS}
\label{sec:methods}

We propose \ours, a one-shot federated conformal calibration strategy designed to handle joint data and model heterogeneity, as shown in Algorithm~\ref{alg:fedwqcp}.

\begin{algorithm}[t]
\caption{One-Shot Federated Conformal Calibration (\ours)}
\label{alg:fedwqcp}
\begin{algorithmic}[1]
\Statex \textbf{Input:} Miscoverage level $\alpha$, $M$ agents with diverse models $\{f_1, \dots, f_M\}$ and calibration sets $\{\mathcal{D}^{\mathrm{cal}}_k\}_{k=1}^M$.
\algcomment{// Agent-side (Local Calibration)}
\For{each agent $k \in \{1,\dots,M\}$ \textbf{in parallel}}
    \State Compute scores $V_{k,i} = V_k(x_{k,i}, y_{k,i})$ for all $(x,y) \in \mathcal{D}^{\mathrm{cal}}_k$
    \State Compute empirical threshold $\widehat q_k = V_{k,(\lceil (n_k+1)(1-\alpha)\rceil)}.$
    \State Send summary $\{(\widehat q_k, n_k)\}_{k=1}^M$ to server
\EndFor

\algcomment{// Server-side (Aggregation)}
\State Receive $\{(\widehat q_k, n_k)\}_{k=1}^M$ and compute aggregated empirical threshold 
\[
\widehat q = \sum_{k=1}^M \frac{n_k}{N}\widehat q_k,
\qquad
N=\sum_{k=1}^M n_k,
\] where $N = \sum_{k=1}^M n_k$.
\State Broadcast $\widehat{q}$ to all agents

\algcomment{// Agent-side (Prediction)}
\State Construct $\mathcal{C}_k(x) = \{ y : V_k(x,y) \le \widehat{q} \}$ for global test inputs.
\Statex \textbf{Output:} $\{\mathcal{C}_k(\cdot)\}_{k=1}^M$
\end{algorithmic}
\end{algorithm}

\subsection{Weighted Quantile Aggregation}

\paragraph{Empirical and Population Quantities.}
For each agent $k \in \{1,\dots,M\}$, let 
$V_{k,1},\dots,V_{k,n_k}$ be calibration nonconformity scores.

Define the empirical CDF: $\hat F_k(v) = \frac{1}{n_k} \sum_{i=1}^{n_k} \mathbf{1}\{V_{k,i} \le v\}$,
and the population CDF: $F_k(v) = \mathbb{P}_{(X,Y)\sim P_k^{\mathrm{cal}}} \big(V_k(X,Y)\le v\big)$.

Define the empirical mixture CDF: $\hat F_{mix}(v) = \sum_{k=1}^M \frac{n_k}{N} \hat F_k(v)$,
and the population mixture CDF: $F_{\mathrm{mix}}(v) = \sum_{k=1}^M \frac{n_k}{N} F_k(v)$,
where $N = \sum_{k=1}^M n_k$.

The local quantile threshold $q_k$ is defined using the 
finite-sample split conformal correction:
\[
q_k = \hat F_k^{-1}(\tau_k),
\quad 
\tau_k = \frac{\lceil (n_k+1)(1-\alpha) \rceil}{n_k}.
\]
Equivalently, $q_k = V_{k,(\lceil (n_k+1)(1-\alpha) \rceil)}$.

In \ours, rather than communicating the full ECDF (which would require sending all $n_k$ scores), agents share only the specific ICDF value $q_k$ and the sample count $n_k$. The server then approximates the global mixture distribution's quantile by aggregating local thresholds via a calibration-size-weighted average: $\hat q = \sum_{k=1}^K \frac{n_k}{N} q_k$.

We emphasize that $\hat q$ is a surrogate for the exact
$(1-\alpha)$ quantile of the pooled score mixture.
In general, because the quantile functional is nonlinear,
$\hat q \neq q_{\mathrm{mix}}$, where $q_{\mathrm{mix}}$
denotes the empirical $(1-\alpha)$ quantile of the mixture
distribution $\hat F_{\mathrm{mix}}$.

\paragraph{Normalization of Heterogeneous Models.} 
Diverse architectures and training intensities ($E$) produce nonconformity scores with different scales and variances. Because the quantile is a rank-based statistic, $q_k$ serves as an architecture-specific normalizer. Transmitting $(q_k, n_k)$ allows the server to synthesize a global boundary that accounts for both predictive strength (value of $q_k$) and statistical reliability (weight $n_k/N$) without ever accessing model parameters.

\subsection{Theoretical Analysis} \label{sec:theory}

We now analyze the coverage behavior of \ours under the dual-heterogeneity setup. Our first result provides a decomposition of the coverage error:

\begin{theorem}[Coverage Decomposition for Surrogate Aggregation]
Let $\hat q$ denote the aggregated threshold and
let $\widehat q_{\mathrm{mix}}= \widehat F_{\mathrm{mix}}^{-1}(1-\alpha)$
denote the empirical mixture quantile.
Then under the target distribution $P_{\mathrm{test}}$,
\begin{flalign*}
&\left|
\mathbb{P}_{P_{\mathrm{test}}}
\!\left(Y \in C_{\widehat q}(X)\right)
- (1-\alpha)
\right| \le
\underbrace{
\left|
\mathbb{P}_{P_{\mathrm{test}}}
\!\left(Y \in C_{\widehat q_{\mathrm{mix}}}(X)\right)
- (1-\alpha)
\right|
}_{\text{calibration-to-test shift term}}
&\\
&\quad+
\underbrace{
\left|
\mathbb{P}_{P_{\mathrm{test}}}
\!\left(Y \in C_{\widehat q}(X)\right)
-
\mathbb{P}_{P_{\mathrm{test}}}
\!\left(Y \in C_{\widehat q_{\mathrm{mix}}}(X)\right)
\right|
}_{\text{aggregation error term}}
&
\end{flalign*}
\end{theorem}

In our design, the first term captures only the effect of calibration-to-test mismatch induced by the Dirichlet split.
No additional training-to-test or cross-agent test shift is present.

\begin{proposition}[Oracle Pooled Split Conformal Threshold]
\label{prop:dist_shift_bound}
Under Assumption~\ref{assump:agent_exchange}, let
$P_{\mathrm{mix}}= \sum_{k=1}^M \frac{n_k}{N} P_k^{\mathrm{cal}}$ with
$N=\sum_{k=1}^M n_k$.
Let $\widehat q_{\mathrm{mix}}^\star$ denote the split conformal threshold
computed from an i.i.d. calibration sample of size $N$ drawn from
$P_{\mathrm{mix}}$ at level $\alpha$.
Then
\begin{flalign*}
&\left|
\mathbb{P}_{P_{\mathrm{test}}}
\!\left(
Y \in C_{\widehat q_{\mathrm{mix}}^\star}(X)
\right)
- (1-\alpha)
\right| \le
\sum_{k=1}^M
\frac{n_k}{N}
\, d_{\mathrm{TV}}
\!\left(
P_k^{\mathrm{cal}},
P_{\mathrm{test}}
\right)
\;+\;\frac{1}{N+1}.
&
\end{flalign*}
\end{proposition}

Proposition~\ref{prop:dist_shift_bound} provides a worst-case upper bound for the ideal pooled quantile $q_{\mathrm{mix}}$ under calibration-to-test shift.
It does not directly imply coverage guarantees for the aggregated threshold $\hat q$, since the latter additionally incurs aggregation error.

\begin{proposition}[A Simple Stability Bound for Quantile Aggregation]
\label{prop:agg_stability}
Let $q_k = F_k^{-1}(1 - \alpha)$ denote the population quantiles and define $q_{\mathrm{avg}} = \sum_{k=1}^M \frac{n_k}{N} q_k$.
Let
\[
F_{mix}(v) = \sum_{k=1}^M \frac{n_k}{N} F_k(v),
\qquad
q_{mix} = F_{mix}^{-1}(1 - \alpha).
\]

Assume there exist constants $\delta > 0$ and $0 < c \le L < \infty$ 
such that for all $v \in [q_{mix} - \delta, q_{mix} + \delta]$,
\[
c \le f_{mix}(v) \le L
\quad \text{and} \quad
0 \le f_k(v) \le L \quad \text{for all } k.
\]

Suppose additionally that the local quantiles satisfy $\max_{1 \le k \le M} |q_k - q_{mix}| \le \delta$.

Then $q_{avg} \in [q_{mix} - \delta, q_{mix} + \delta]$ and
\[
|q_{avg} - q_{mix}|
\le \frac{L}{c}
\sum_{j=1}^M \sum_{k=1}^M
\frac{n_j}{N} \frac{n_k}{N}
|q_j - q_k|.
\]
\end{proposition}

\paragraph{Proof (sketch).}
By the inverse function theorem for monotone CDFs and the lower density bound
$f_{\mathrm{mix}}\ge c$ near $q_{\mathrm{mix}}$,
\[
|q_{\mathrm{avg}}-q_{\mathrm{mix}}|
\le \frac{1}{c}\,\big|F_{\mathrm{mix}}(q_{\mathrm{avg}}) - F_{\mathrm{mix}}(q_{\mathrm{mix}})\big|
= \frac{1}{c}\,\big|F_{\mathrm{mix}}(q_{\mathrm{avg}}) - (1-\alpha)\big|.
\]
Next,
\[
F_{\mathrm{mix}}(q_{\mathrm{avg}})-(1-\alpha)
=
\sum_{k=1}^M \frac{n_k}{N}\big(F_k(q_{\mathrm{avg}})-F_k(q_k)\big),
\]
and by the upper density bound $f_k\le L$,
\[
|F_k(q_{\mathrm{avg}})-F_k(q_k)| \le L|q_{\mathrm{avg}}-q_k|.
\]
Therefore
\begin{flalign*}
&\left|
F_{\mathrm{mix}}(q_{\mathrm{avg}})
-
(1-\alpha)
\right|
\le
L\sum_{k=1}^M \frac{n_k}{N}|q_{\mathrm{avg}}-q_k |
&\\
&=
L\sum_{k=1}^M \frac{n_k}{N}
\Big|
\sum_{j=1}^M\frac{n_j}{N}(q_j-q_k)
\Big|
\le
L\sum_{j=1}^M \sum_{k=1}^M \frac{n_j}{N}\frac{n_k}{N}|q_j-q_k|.
&
\end{flalign*}
Combining yields the claim.

\paragraph{Regularity assumption.}
Proposition~\ref{prop:agg_stability} is population-level and assumes local differentiability with densities bounded as stated; discrete empirical scores may violate these conditions.

\paragraph{Clarification on aggregation bias.}
The condition $|\widehat q-q_{\mathrm{mix}}|\to 0$ is not automatic since quantiles are nonlinear; it holds, for example, when client score distributions become asymptotically aligned.

\paragraph{Remark 1.}
Our empirical results in Section~\ref{sec:empirical_res} and Section~\ref{sec:ablation} suggest that this aggregation-induced deviation does not dominate coverage behavior in the heterogeneous regimes considered.
Proofs of Proposition~\ref{prop:dist_shift_bound} and Proposition~\ref{prop:agg_stability} are provided in Appendix~\ref{app:proof_shift_bound} and Appendix~\ref{app:proof_prop_2}, respectively.

\paragraph{Remark 2.}
The algorithm uses empirical thresholds $\widehat q_k$
and $\widehat q=\sum (n_k/N)\widehat q_k$.
Under uniform convergence of $\widehat F_k$ to $F_k$
(as $n_k\to\infty$),
we have $\widehat q_k\to q_k$ and hence
$\widehat q\to q_{\mathrm{avg}}$.
Thus Proposition~\ref{prop:agg_stability} characterizes the limiting aggregation bias.

\begin{theorem}[Asymptotic Behavior Under Vanishing Calibration Shift]
Consider a joint asymptotic regime in which
\begin{enumerate}[label=(\roman*)]
  \item $n_k \to \infty$ for all $k$,
  \item the Dirichlet concentration parameter $\beta \to \infty$,
  \item the number of agents $M$ is fixed.
\end{enumerate}
Assume that under this regime
\[
\sup_{k} d_{TV}\!\left(P_k^{\mathrm{cal}}, P_{\mathrm{test}}\right) \to 0,
\qquad\text{and}\qquad
\big|\widehat q - q_{\mathrm{mix}}\big| \to 0.
\]
Then $\left| \mathbb{P}_{P_{\mathrm{test}}}\!\left(Y \in C_{\widehat q}(X)\right) - (1-\alpha) \right| \to 0$.
\end{theorem}

\paragraph{Remark 3.}
This result is asymptotic and does not imply finite-sample coverage guarantees under heterogeneous calibration splits.

\paragraph{Remark 4.}
The condition $|\widehat q - q_{\mathrm{mix}}| \to 0$ is not automatic.
In general, averaging quantiles does not coincide with the quantile
of a mixture distribution, since the quantile functional is nonlinear.
Convergence of the aggregation bias requires that the client score
distributions become asymptotically aligned, for example through
vanishing calibration heterogeneity (e.g., $\beta \to \infty$)
or through convergence of the population quantiles
$q_k \to q_{\mathrm{mix}}$ as $n_k \to \infty$.
The theorem should therefore be interpreted as describing a
regime in which both distributional heterogeneity and
quantile dispersion diminish.
The theorem should be interpreted as describing an asymptotic regime in which both distributional heterogeneity and aggregation bias vanish.
\section{EMPIRICAL EVALUATION} \label{sec:empirical_res}
\textbf{Benchmarks and Data Partitioning.} We evaluate \ours across seven public datasets most frequently used by prior federated UQ works~\cite{lu2023federated,minpersonalized}, ensuring a fair and comprehensive comparison with established baselines. This suite includes three standard vision benchmarks (MNIST~\cite{lecun1998mnist}, FashionMNIST~\cite{xiao2017fashion}, and CIFAR-10~\cite{krizhevsky2009learning}) and four specialized medical imaging datasets (DermaMNIST~\cite{tschandl2018ham10000}, BloodMNIST~\cite{acevedo2020dataset}, TissueMNIST~\cite{ljosa2012annotated}, and RetinaMNIST~\cite{dataset20202nd}). To simulate realistic data heterogeneity, we apply a Dirichlet partition to induce label shift in classification and covariate shift in regression tasks. Detailed data statistics and partition descriptions are provided in Appendix~\ref{app:dat_stats}.

\textbf{Experimental Design for Dual Heterogeneity.} To rigorously evaluate \ours under joint heterogeneity, we simulate a FL system comprising six agents with divergent predictive strengths. Following the setup in \cite{wang2025asymmetrical}, we designate three strong agents and three weak agents. These agents utilize heterogeneous architectures to reflect the hardware and training disparities common in cross-silo FL. Further details on the model architectures and the controlled heterogeneity setup are available in Appendix~\ref{app:study_design}.

\subsection{Marginal Coverage Across Datasets}
Table~\ref{tab:all_datasets_cov} reports the empirical marginal coverage at both agent and global levels across seven datasets. The results validate our theoretical framework through two key observations. 
First, \ours empirically maintains coverage around the nominal level at both agent and global levels across all datasets. In contrast, existing federated UQ baselines either under-cover or exhibit systematic over-coverage, with DP-FedCP in particular consistently exhibiting severe under-coverage across datasets. This validates the robustness of weighted quantile aggregation under joint data and model heterogeneity. 
Second, \ours maintains a competitive runtime across all benchmarks, notably outperforming iterative or heavy-pooling methods like CPhet and DP-FedCP. This validates the efficiency of the one-shot procedure, demonstrating that \ours provides strong empirical reliability with minimal communication overhead of transmitting only two scalars per agent.

\begin{table*}[!t]
\centering
\caption{
Empirical marginal coverage across seven benchmarks under dual heterogeneity. 
We report the agent-level coverage for strong (\textbf{S0--S2}) and weak (\textbf{W3--W5}) agents, alongside the global average (\textbf{Avg}). Results represent the median $\pm$ 95\% CI over 10 independent runs (target error $\alpha = 0.05$, partition $\text{Dir}(0.3)$). All runtimes are measured in seconds. \ours empirically maintains coverage around the nominal level across all datasets with a one-shot runtime comparable to the most efficient baselines.
}
\label{tab:all_datasets_cov}
\resizebox{\textwidth}{!}{
\begin{tabular}{llcccccccc}
\toprule
\textbf{Dataset} & \textbf{Method} 
& \textbf{S0} & \textbf{S1} & \textbf{S2} 
& \textbf{W3} & \textbf{W4} & \textbf{W5} 
& \textbf{Avg} & \textbf{Runtime (s)} \\
\midrule

\multirow{6}{*}{MNIST}

& DP-FedCP 
& $0.0610_{\scriptscriptstyle \pm 0.0131}$
& $0.0668_{\scriptscriptstyle \pm 0.0109}$
& $0.0675_{\scriptscriptstyle \pm 0.0146}$
& $0.7251_{\scriptscriptstyle \pm 0.0319}$
& $0.7214_{\scriptscriptstyle \pm 0.0317}$
& $0.7191_{\scriptscriptstyle \pm 0.0258}$
& $0.3957_{\scriptscriptstyle \pm 0.0033}$
& $6.383_{\scriptscriptstyle \pm 0.109}$ \\

& SplitCP 
& $1.0000_{\scriptscriptstyle \pm 0.0000}$
& $1.0000_{\scriptscriptstyle \pm 0.0000}$
& $1.0000_{\scriptscriptstyle \pm 0.0000}$
& $0.9965_{\scriptscriptstyle \pm 0.0019}$
& $0.9960_{\scriptscriptstyle \pm 0.0024}$
& $0.9963_{\scriptscriptstyle \pm 0.0023}$
& $0.9981_{\scriptscriptstyle \pm 0.0004}$
& -- \\

& FedCP-QQ 
& $1.0000_{\scriptscriptstyle \pm 0.0000}$
& $1.0000_{\scriptscriptstyle \pm 0.0000}$
& $1.0000_{\scriptscriptstyle \pm 0.0000}$
& $1.0000_{\scriptscriptstyle \pm 0.0001}$
& $1.0000_{\scriptscriptstyle \pm 0.0000}$
& $1.0000_{\scriptscriptstyle \pm 0.0001}$
& $1.0000_{\scriptscriptstyle \pm 0.0000}$
& $2.887_{\scriptscriptstyle \pm 0.027}$ \\

& FCP 
& $1.0000_{\scriptscriptstyle \pm 0.0000}$
& $1.0000_{\scriptscriptstyle \pm 0.0000}$
& $1.0000_{\scriptscriptstyle \pm 0.0000}$
& $0.9999_{\scriptscriptstyle \pm 0.0001}$
& $0.9999_{\scriptscriptstyle \pm 0.0001}$
& $0.9999_{\scriptscriptstyle \pm 0.0001}$
& $0.9999_{\scriptscriptstyle \pm 0.0001}$
& $3.817_{\scriptscriptstyle \pm 0.124}$ \\

& CPhet 
& $1.0000_{\scriptscriptstyle \pm 0.0000}$
& $1.0000_{\scriptscriptstyle \pm 0.0000}$
& $1.0000_{\scriptscriptstyle \pm 0.0000}$
& $0.9967_{\scriptscriptstyle \pm 0.0017}$
& $0.9961_{\scriptscriptstyle \pm 0.0025}$
& $0.9966_{\scriptscriptstyle \pm 0.0024}$
& $0.9982_{\scriptscriptstyle \pm 0.0004}$
& $5.955_{\scriptscriptstyle \pm 0.061}$ \\
\cmidrule(lr){2-10}
& \cellcolor{lightblue}\textbf{\ours}
& \cellcolor{lightblue}$\mathbf{0.9978_{\scriptscriptstyle \pm 0.0012}}$
& \cellcolor{lightblue}$\mathbf{0.9979_{\scriptscriptstyle \pm 0.0012}}$
& \cellcolor{lightblue}$\mathbf{0.9976_{\scriptscriptstyle \pm 0.0011}}$
& \cellcolor{lightblue}$\mathbf{0.9977_{\scriptscriptstyle \pm 0.0010}}$
& \cellcolor{lightblue}$\mathbf{0.9973_{\scriptscriptstyle \pm 0.0010}}$
& \cellcolor{lightblue}$\mathbf{0.9976_{\scriptscriptstyle \pm 0.0012}}$
& \cellcolor{lightblue}$\mathbf{0.9977_{\scriptscriptstyle \pm 0.0009}}$
& \cellcolor{lightblue}$\mathbf{2.831_{\scriptscriptstyle \pm 0.105}}$ \\

\midrule

\multirow{6}{*}{FashionMNIST}

& DP-FedCP 
& $0.3831_{\scriptscriptstyle \pm 0.0504}$
& $0.3841_{\scriptscriptstyle \pm 0.0360}$
& $0.3859_{\scriptscriptstyle \pm 0.0277}$
& $0.7276_{\scriptscriptstyle \pm 0.0366}$
& $0.7379_{\scriptscriptstyle \pm 0.0223}$
& $0.7404_{\scriptscriptstyle \pm 0.0214}$
& $0.5589_{\scriptscriptstyle \pm 0.0067}$
& $5.893_{\scriptscriptstyle \pm 0.102}$ \\

& SplitCP 
& $0.9998_{\scriptscriptstyle \pm 0.0025}$
& $0.9999_{\scriptscriptstyle \pm 0.0018}$
& $0.9998_{\scriptscriptstyle \pm 0.0008}$
& $0.9928_{\scriptscriptstyle \pm 0.0077}$
& $0.9975_{\scriptscriptstyle \pm 0.0037}$
& $0.9962_{\scriptscriptstyle \pm 0.0081}$
& $0.9967_{\scriptscriptstyle \pm 0.0016}$
& -- \\

& FedCP-QQ 
& $1.0000_{\scriptscriptstyle \pm 0.0002}$
& $1.0000_{\scriptscriptstyle \pm 0.0004}$
& $1.0000_{\scriptscriptstyle \pm 0.0001}$
& $1.0000_{\scriptscriptstyle \pm 0.0002}$
& $1.0000_{\scriptscriptstyle \pm 0.0002}$
& $1.0000_{\scriptscriptstyle \pm 0.0002}$
& $1.0000_{\scriptscriptstyle \pm 0.0002}$
& $2.638_{\scriptscriptstyle \pm 0.119}$ \\

& FCP 
& $0.9995_{\scriptscriptstyle \pm 0.0005}$
& $0.9997_{\scriptscriptstyle \pm 0.0006}$
& $0.9995_{\scriptscriptstyle \pm 0.0008}$
& $0.9997_{\scriptscriptstyle \pm 0.0001}$
& $0.9997_{\scriptscriptstyle \pm 0.0001}$
& $0.9997_{\scriptscriptstyle \pm 0.0002}$
& $0.9996_{\scriptscriptstyle \pm 0.0003}$
& $3.511_{\scriptscriptstyle \pm 0.095}$ \\

& CPhet 
& $0.9998_{\scriptscriptstyle \pm 0.0025}$
& $0.9999_{\scriptscriptstyle \pm 0.0018}$
& $0.9998_{\scriptscriptstyle \pm 0.0008}$
& $0.9909_{\scriptscriptstyle \pm 0.0097}$
& $0.9974_{\scriptscriptstyle \pm 0.0034}$
& $0.9956_{\scriptscriptstyle \pm 0.0066}$
& $0.9967_{\scriptscriptstyle \pm 0.0015}$
& $5.409_{\scriptscriptstyle \pm 0.122}$ \\
\cmidrule(lr){2-10}
& \cellcolor{lightblue}\textbf{\ours}
& \cellcolor{lightblue}$\mathbf{0.9906_{\scriptscriptstyle \pm 0.0060}}$
& \cellcolor{lightblue}$\mathbf{0.9919_{\scriptscriptstyle \pm 0.0036}}$
& \cellcolor{lightblue}$\mathbf{0.9899_{\scriptscriptstyle \pm 0.0046}}$
& \cellcolor{lightblue}$\mathbf{0.9973_{\scriptscriptstyle \pm 0.0017}}$
& \cellcolor{lightblue}$\mathbf{0.9975_{\scriptscriptstyle \pm 0.0016}}$
& \cellcolor{lightblue}$\mathbf{0.9971_{\scriptscriptstyle \pm 0.0018}}$
& \cellcolor{lightblue}$\mathbf{0.9942_{\scriptscriptstyle \pm 0.0028}}$
& \cellcolor{lightblue}$\mathbf{2.636_{\scriptscriptstyle \pm 0.056}}$ \\

\midrule

\multirow{6}{*}{CIFAR-10}

& DP-FedCP 
& $0.7573_{\scriptscriptstyle \pm 0.0308}$
& $0.7334_{\scriptscriptstyle \pm 0.0261}$
& $0.7508_{\scriptscriptstyle \pm 0.0345}$
& $0.9149_{\scriptscriptstyle \pm 0.0117}$
& $0.9136_{\scriptscriptstyle \pm 0.0090}$
& $0.9152_{\scriptscriptstyle \pm 0.0150}$
& $0.8300_{\scriptscriptstyle \pm 0.0082}$
& $5.371_{\scriptscriptstyle \pm 0.158}$ \\

& SplitCP 
& $0.9829_{\scriptscriptstyle \pm 0.0081}$
& $0.9878_{\scriptscriptstyle \pm 0.0160}$
& $0.9829_{\scriptscriptstyle \pm 0.0114}$
& $0.9550_{\scriptscriptstyle \pm 0.0191}$
& $0.9419_{\scriptscriptstyle \pm 0.0107}$
& $0.9479_{\scriptscriptstyle \pm 0.0154}$
& $0.9666_{\scriptscriptstyle \pm 0.0045}$
& -- \\

& FedCP-QQ 
& $1.0000_{\scriptscriptstyle \pm 0.0089}$
& $1.0000_{\scriptscriptstyle \pm 0.0108}$
& $1.0000_{\scriptscriptstyle \pm 0.0106}$
& $1.0000_{\scriptscriptstyle \pm 0.0157}$
& $1.0000_{\scriptscriptstyle \pm 0.0146}$
& $1.0000_{\scriptscriptstyle \pm 0.0155}$
& $1.0000_{\scriptscriptstyle \pm 0.0125}$
& $2.353_{\scriptscriptstyle \pm 0.031}$ \\

& FCP 
& $0.9809_{\scriptscriptstyle \pm 0.0104}$
& $0.9787_{\scriptscriptstyle \pm 0.0075}$
& $0.9806_{\scriptscriptstyle \pm 0.0065}$
& $0.9687_{\scriptscriptstyle \pm 0.0037}$
& $0.9711_{\scriptscriptstyle \pm 0.0060}$
& $0.9699_{\scriptscriptstyle \pm 0.0045}$
& $0.9750_{\scriptscriptstyle \pm 0.0054}$
& $3.159_{\scriptscriptstyle \pm 0.067}$ \\

& CPhet 
& $0.9839_{\scriptscriptstyle \pm 0.0063}$
& $0.9872_{\scriptscriptstyle \pm 0.0155}$
& $0.9842_{\scriptscriptstyle \pm 0.0087}$
& $0.9366_{\scriptscriptstyle \pm 0.0140}$
& $0.9413_{\scriptscriptstyle \pm 0.0123}$
& $0.9395_{\scriptscriptstyle \pm 0.0029}$
& $0.9623_{\scriptscriptstyle \pm 0.0038}$
& $4.875_{\scriptscriptstyle \pm 0.070}$ \\
\cmidrule(lr){2-10}
& \cellcolor{lightblue}\textbf{\ours}
& \cellcolor{lightblue}$\mathbf{0.9657_{\scriptscriptstyle \pm 0.0095}}$
& \cellcolor{lightblue}$\mathbf{0.9625_{\scriptscriptstyle \pm 0.0088}}$
& \cellcolor{lightblue}$\mathbf{0.9653_{\scriptscriptstyle \pm 0.0067}}$
& \cellcolor{lightblue}$\mathbf{0.9575_{\scriptscriptstyle \pm 0.0046}}$
& \cellcolor{lightblue}$\mathbf{0.9609_{\scriptscriptstyle \pm 0.0069}}$
& \cellcolor{lightblue}$\mathbf{0.9594_{\scriptscriptstyle \pm 0.0043}}$
& \cellcolor{lightblue}$\mathbf{0.9621_{\scriptscriptstyle \pm 0.0045}}$
& \cellcolor{lightblue}$\mathbf{2.366_{\scriptscriptstyle \pm 0.039}}$ \\

\midrule

\multirow{6}{*}{DermaMNIST}

& DP-FedCP 
& $0.7257_{\scriptscriptstyle \pm 0.0998}$
& $0.7382_{\scriptscriptstyle \pm 0.0701}$
& $0.6970_{\scriptscriptstyle \pm 0.0818}$
& $0.9212_{\scriptscriptstyle \pm 0.0436}$
& $0.8975_{\scriptscriptstyle \pm 0.0750}$
& $0.8793_{\scriptscriptstyle \pm 0.0774}$
& $0.8107_{\scriptscriptstyle \pm 0.0388}$
& $0.953_{\scriptscriptstyle \pm 0.016}$ \\

& SplitCP 
& $0.9890_{\scriptscriptstyle \pm 0.0143}$
& $0.9853_{\scriptscriptstyle \pm 0.0128}$
& $0.9923_{\scriptscriptstyle \pm 0.0092}$
& $0.9688_{\scriptscriptstyle \pm 0.0278}$
& $0.9656_{\scriptscriptstyle \pm 0.0189}$
& $0.9661_{\scriptscriptstyle \pm 0.0145}$
& $0.9756_{\scriptscriptstyle \pm 0.0083}$
& -- \\

& FedCP-QQ 
& $0.9998_{\scriptscriptstyle \pm 0.0040}$
& $0.9995_{\scriptscriptstyle \pm 0.0046}$
& $0.9998_{\scriptscriptstyle \pm 0.0037}$
& $0.9988_{\scriptscriptstyle \pm 0.0073}$
& $0.9988_{\scriptscriptstyle \pm 0.0074}$
& $0.9993_{\scriptscriptstyle \pm 0.0069}$
& $0.9993_{\scriptscriptstyle \pm 0.0055}$
& $0.311_{\scriptscriptstyle \pm 0.012}$ \\

& FCP 
& $0.9880_{\scriptscriptstyle \pm 0.0111}$
& $0.9813_{\scriptscriptstyle \pm 0.0098}$
& $0.9843_{\scriptscriptstyle \pm 0.0127}$
& $0.9823_{\scriptscriptstyle \pm 0.0126}$
& $0.9813_{\scriptscriptstyle \pm 0.0112}$
& $0.9800_{\scriptscriptstyle \pm 0.0202}$
& $0.9838_{\scriptscriptstyle \pm 0.0100}$
& $0.475_{\scriptscriptstyle \pm 0.060}$ \\

& CPhet 
& $0.9873_{\scriptscriptstyle \pm 0.0289}$
& $0.9803_{\scriptscriptstyle \pm 0.0191}$
& $0.9885_{\scriptscriptstyle \pm 0.0244}$
& $0.9628_{\scriptscriptstyle \pm 0.0166}$
& $0.9544_{\scriptscriptstyle \pm 0.0103}$
& $0.9603_{\scriptscriptstyle \pm 0.0143}$
& $0.9704_{\scriptscriptstyle \pm 0.0077}$
& $0.667_{\scriptscriptstyle \pm 0.015}$ \\
\cmidrule(lr){2-10}
& \cellcolor{lightblue}\textbf{\ours}
& \cellcolor{lightblue}$\mathbf{0.9788_{\scriptscriptstyle \pm 0.0209}}$
& \cellcolor{lightblue}$\mathbf{0.9708_{\scriptscriptstyle \pm 0.0149}}$
& \cellcolor{lightblue}$\mathbf{0.9703_{\scriptscriptstyle \pm 0.0242}}$
& \cellcolor{lightblue}$\mathbf{0.9683_{\scriptscriptstyle \pm 0.0158}}$
& \cellcolor{lightblue}$\mathbf{0.9726_{\scriptscriptstyle \pm 0.0170}}$
& \cellcolor{lightblue}$\mathbf{0.9731_{\scriptscriptstyle \pm 0.0209}}$
& \cellcolor{lightblue}$\mathbf{0.9680_{\scriptscriptstyle \pm 0.0161}}$
& \cellcolor{lightblue}$\mathbf{0.304_{\scriptscriptstyle \pm 0.014}}$ \\

\midrule

\multirow{6}{*}{BloodMNIST}

& DP-FedCP 
& $0.4803_{\scriptscriptstyle \pm 0.0642}$
& $0.4911_{\scriptscriptstyle \pm 0.0563}$
& $0.5020_{\scriptscriptstyle \pm 0.0607}$
& $0.8499_{\scriptscriptstyle \pm 0.0428}$
& $0.8582_{\scriptscriptstyle \pm 0.0187}$
& $0.8642_{\scriptscriptstyle \pm 0.0400}$
& $0.6800_{\scriptscriptstyle \pm 0.0123}$
& $5.805_{\scriptscriptstyle \pm 0.480}$ \\

& SplitCP 
& $0.9994_{\scriptscriptstyle \pm 0.0022}$
& $0.9994_{\scriptscriptstyle \pm 0.0009}$
& $0.9994_{\scriptscriptstyle \pm 0.0034}$
& $0.9671_{\scriptscriptstyle \pm 0.0267}$
& $0.9737_{\scriptscriptstyle \pm 0.0156}$
& $0.9752_{\scriptscriptstyle \pm 0.0071}$
& $0.9842_{\scriptscriptstyle \pm 0.0045}$
& -- \\

& FedCP-QQ 
& $1.0000_{\scriptscriptstyle \pm 0.0023}$
& $1.0000_{\scriptscriptstyle \pm 0.0021}$
& $1.0000_{\scriptscriptstyle \pm 0.0015}$
& $0.9994_{\scriptscriptstyle \pm 0.0059}$
& $1.0000_{\scriptscriptstyle \pm 0.0050}$
& $0.9994_{\scriptscriptstyle \pm 0.0056}$
& $0.9997_{\scriptscriptstyle \pm 0.0035}$
& $0.578_{\scriptscriptstyle \pm 0.027}$ \\

& FCP 
& $0.9993_{\scriptscriptstyle \pm 0.0008}$
& $0.9996_{\scriptscriptstyle \pm 0.0004}$
& $0.9994_{\scriptscriptstyle \pm 0.0013}$
& $0.9959_{\scriptscriptstyle \pm 0.0019}$
& $0.9969_{\scriptscriptstyle \pm 0.0033}$
& $0.9947_{\scriptscriptstyle \pm 0.0018}$
& $0.9976_{\scriptscriptstyle \pm 0.0012}$
& $0.893_{\scriptscriptstyle \pm 0.049}$ \\

& CPhet 
& $0.9994_{\scriptscriptstyle \pm 0.0014}$
& $0.9994_{\scriptscriptstyle \pm 0.0009}$
& $0.9994_{\scriptscriptstyle \pm 0.0035}$
& $0.9671_{\scriptscriptstyle \pm 0.0214}$
& $0.9735_{\scriptscriptstyle \pm 0.0134}$
& $0.9759_{\scriptscriptstyle \pm 0.0101}$
& $0.9850_{\scriptscriptstyle \pm 0.0049}$
& $1.231_{\scriptscriptstyle \pm 0.078}$ \\
\cmidrule(lr){2-10}
& \cellcolor{lightblue}\textbf{\ours}
& \cellcolor{lightblue}$\mathbf{0.9868_{\scriptscriptstyle \pm 0.0073}}$
& \cellcolor{lightblue}$\mathbf{0.9868_{\scriptscriptstyle \pm 0.0070}}$
& \cellcolor{lightblue}$\mathbf{0.9886_{\scriptscriptstyle \pm 0.0087}}$
& \cellcolor{lightblue}$\mathbf{0.9775_{\scriptscriptstyle \pm 0.0047}}$
& \cellcolor{lightblue}$\mathbf{0.9794_{\scriptscriptstyle \pm 0.0073}}$
& \cellcolor{lightblue}$\mathbf{0.9768_{\scriptscriptstyle \pm 0.0059}}$
& \cellcolor{lightblue}$\mathbf{0.9824_{\scriptscriptstyle \pm 0.0062}}$
& \cellcolor{lightblue}$\mathbf{0.571_{\scriptscriptstyle \pm 0.040}}$ \\

\midrule

\multirow{6}{*}{TissueMNIST}

& DP-FedCP 
& $0.8485_{\scriptscriptstyle \pm 0.0470}$
& $0.8547_{\scriptscriptstyle \pm 0.0429}$
& $0.8611_{\scriptscriptstyle \pm 0.0514}$
& $0.9431_{\scriptscriptstyle \pm 0.0298}$
& $0.9433_{\scriptscriptstyle \pm 0.0207}$
& $0.9389_{\scriptscriptstyle \pm 0.0429}$
& $0.8951_{\scriptscriptstyle \pm 0.0207}$
& $13.293_{\scriptscriptstyle \pm 0.598}$ \\

& SplitCP 
& $0.9646_{\scriptscriptstyle \pm 0.0201}$
& $0.9635_{\scriptscriptstyle \pm 0.0173}$
& $0.9542_{\scriptscriptstyle \pm 0.0203}$
& $0.9646_{\scriptscriptstyle \pm 0.0567}$
& $0.9621_{\scriptscriptstyle \pm 0.0277}$
& $0.9565_{\scriptscriptstyle \pm 0.0373}$
& $0.9576_{\scriptscriptstyle \pm 0.0084}$
& -- \\

& FedCP-QQ 
& $1.0000_{\scriptscriptstyle \pm 0.0248}$
& $1.0000_{\scriptscriptstyle \pm 0.0255}$
& $1.0000_{\scriptscriptstyle \pm 0.0252}$
& $1.0000_{\scriptscriptstyle \pm 0.0213}$
& $1.0000_{\scriptscriptstyle \pm 0.0226}$
& $1.0000_{\scriptscriptstyle \pm 0.0236}$
& $1.0000_{\scriptscriptstyle \pm 0.0238}$
& $6.364_{\scriptscriptstyle \pm 0.370}$ \\

& FCP 
& $0.9572_{\scriptscriptstyle \pm 0.0091}$
& $0.9602_{\scriptscriptstyle \pm 0.0109}$
& $0.9623_{\scriptscriptstyle \pm 0.0121}$
& $0.9644_{\scriptscriptstyle \pm 0.0068}$
& $0.9649_{\scriptscriptstyle \pm 0.0071}$
& $0.9643_{\scriptscriptstyle \pm 0.0086}$
& $0.9634_{\scriptscriptstyle \pm 0.0076}$
& $8.463_{\scriptscriptstyle \pm 0.454}$ \\

& CPhet 
& $0.9736_{\scriptscriptstyle \pm 0.0189}$
& $0.9657_{\scriptscriptstyle \pm 0.0144}$
& $0.9559_{\scriptscriptstyle \pm 0.0204}$
& $0.9515_{\scriptscriptstyle \pm 0.0264}$
& $0.9529_{\scriptscriptstyle \pm 0.0211}$
& $0.9443_{\scriptscriptstyle \pm 0.0213}$
& $0.9574_{\scriptscriptstyle \pm 0.0059}$
& $13.073_{\scriptscriptstyle \pm 0.418}$ \\
\cmidrule(lr){2-10}
& \cellcolor{lightblue}\textbf{\ours}
& \cellcolor{lightblue}$\mathbf{0.9408_{\scriptscriptstyle \pm 0.0164}}$
& \cellcolor{lightblue}$\mathbf{0.9459_{\scriptscriptstyle \pm 0.0211}}$
& \cellcolor{lightblue}$\mathbf{0.9471_{\scriptscriptstyle \pm 0.0190}}$
& \cellcolor{lightblue}$\mathbf{0.9589_{\scriptscriptstyle \pm 0.0135}}$
& \cellcolor{lightblue}$\mathbf{0.9562_{\scriptscriptstyle \pm 0.0117}}$
& \cellcolor{lightblue}$\mathbf{0.9547_{\scriptscriptstyle \pm 0.0141}}$
& \cellcolor{lightblue}$\mathbf{0.9499_{\scriptscriptstyle \pm 0.0151}}$
& \cellcolor{lightblue}$\mathbf{6.413_{\scriptscriptstyle \pm 0.320}}$ \\

\midrule

\multirow{6}{*}{RetinaMNIST}

& SplitCP 
& $0.9612_{\scriptscriptstyle \pm 0.1018}$
& $0.9500_{\scriptscriptstyle \pm 0.1286}$
& $0.9750_{\scriptscriptstyle \pm 0.0530}$
& $0.9125_{\scriptscriptstyle \pm 0.1213}$
& $0.9338_{\scriptscriptstyle \pm 0.0867}$
& $0.9738_{\scriptscriptstyle \pm 0.0330}$
& $0.9417_{\scriptscriptstyle \pm 0.0313}$
& -- \\

& FedCP-QQ 
& $0.9850_{\scriptscriptstyle \pm 0.0766}$
& $0.9850_{\scriptscriptstyle \pm 0.0686}$
& $0.9850_{\scriptscriptstyle \pm 0.0706}$
& $0.9788_{\scriptscriptstyle \pm 0.1238}$
& $0.9863_{\scriptscriptstyle \pm 0.0845}$
& $0.9812_{\scriptscriptstyle \pm 0.0718}$
& $0.9823_{\scriptscriptstyle \pm 0.0824}$
& $0.038_{\scriptscriptstyle \pm 0.004}$ \\

& FCP 
& $0.9812_{\scriptscriptstyle \pm 0.0050}$
& $0.9800_{\scriptscriptstyle \pm 0.0085}$
& $0.9812_{\scriptscriptstyle \pm 0.0079}$
& $0.9663_{\scriptscriptstyle \pm 0.0258}$
& $0.9675_{\scriptscriptstyle \pm 0.0269}$
& $0.9775_{\scriptscriptstyle \pm 0.0121}$
& $0.9735_{\scriptscriptstyle \pm 0.0078}$
& $0.052_{\scriptscriptstyle \pm 0.003}$ \\

& CPhet 
& $0.9675_{\scriptscriptstyle \pm 0.0175}$
& $0.9688_{\scriptscriptstyle \pm 0.0157}$
& $0.9738_{\scriptscriptstyle \pm 0.0173}$
& $0.9550_{\scriptscriptstyle \pm 0.0359}$
& $0.9500_{\scriptscriptstyle \pm 0.0295}$
& $0.9550_{\scriptscriptstyle \pm 0.0146}$
& $0.9592_{\scriptscriptstyle \pm 0.0152}$
& $0.077_{\scriptscriptstyle \pm 0.002}$ \\
\cmidrule(lr){2-10}
& \cellcolor{lightblue}\textbf{\ours}
& \cellcolor{lightblue}$\mathbf{0.9625_{\scriptscriptstyle \pm 0.0183}}$
& \cellcolor{lightblue}$\mathbf{0.9637_{\scriptscriptstyle \pm 0.0177}}$
& \cellcolor{lightblue}$\mathbf{0.9575_{\scriptscriptstyle \pm 0.0186}}$
& \cellcolor{lightblue}$\mathbf{0.9425_{\scriptscriptstyle \pm 0.0308}}$
& \cellcolor{lightblue}$\mathbf{0.9500_{\scriptscriptstyle \pm 0.0291}}$
& \cellcolor{lightblue}$\mathbf{0.9550_{\scriptscriptstyle \pm 0.0374}}$
& \cellcolor{lightblue}$\mathbf{0.9542_{\scriptscriptstyle \pm 0.0197}}$
& \cellcolor{lightblue}$\mathbf{0.034_{\scriptscriptstyle \pm 0.001}}$ \\

\bottomrule

\end{tabular}
}
\end{table*}

\subsection{Efficiency Comparison at Target Coverage}

\begin{table}[H]
\centering
\caption{
Empirical inefficiency measured by average prediction set size (classification) or interval length (regression). Smaller is better.
Parentheses show percentage reduction relative to \ours. Values are median $\pm$ 95\% CI over 10 runs. \ours consistently 
achieves lower inefficiency, producing substantially smaller prediction sets or intervals compared to existing federated UQ baselines.
}
\label{tab:efficiency_comparison}

\setlength{\tabcolsep}{3.5pt}
\renewcommand{\arraystretch}{1.02}

\small
\resizebox{0.55\textwidth}{!}{
\begin{tabular}{lcc}
\toprule
\textbf{Dataset} & \textbf{Method} & \textbf{Avg Size} (\% $\downarrow$) \\
\midrule

\multirow{5}{*}{MNIST}
& SplitCP  & $6.88_{\scriptscriptstyle \pm 0.64}(56.4\%)$ \\
& FedCP-QQ & $9.59_{\scriptscriptstyle \pm 0.21}(68.7\%)$ \\
& FCP      & $7.28_{\scriptscriptstyle \pm 1.56}(58.8\%)$ \\
& CPhet    & $6.90_{\scriptscriptstyle \pm 0.65}(56.5\%)$ \\
\cmidrule(lr){2-3}
& \cellcolor{lightblue}\textbf{\ours}
& \cellcolor{lightblue}$\mathbf{3.00_{\scriptscriptstyle \pm 0.24}}$ \\

\midrule

\multirow{5}{*}{FashionMNIST}
& SplitCP  & $4.68_{\scriptscriptstyle \pm 0.82}(25.4\%)$ \\
& FedCP-QQ & $9.68_{\scriptscriptstyle \pm 2.15}(63.9\%)$ \\
& FCP      & $5.60_{\scriptscriptstyle \pm 0.90}(37.7\%)$ \\
& CPhet    & $4.65_{\scriptscriptstyle \pm 0.85}(24.9\%)$ \\
\cmidrule(lr){2-3}
& \cellcolor{lightblue}\textbf{\ours}
& \cellcolor{lightblue}$\mathbf{3.49_{\scriptscriptstyle \pm 0.34}}$ \\

\midrule

\multirow{4}{*}{CIFAR-10}
& SplitCP  & $6.17_{\scriptscriptstyle \pm 0.24}(5.7\%)$ \\
& FedCP-QQ & $10.00_{\scriptscriptstyle \pm 1.77}(41.8\%)$ \\
& FCP      & $6.46_{\scriptscriptstyle \pm 0.32}(9.9\%)$ \\
\cmidrule(lr){2-3}
& \cellcolor{lightblue}\textbf{\ours}
& \cellcolor{lightblue}$\mathbf{5.82_{\scriptscriptstyle \pm 0.20}}$ \\

\midrule

\multirow{5}{*}{DermaMNIST}
& SplitCP  & $3.90_{\scriptscriptstyle \pm 0.41}(9.7\%)$ \\
& FedCP-QQ & $6.85_{\scriptscriptstyle \pm 1.06}(48.6\%)$ \\
& FCP      & $4.18_{\scriptscriptstyle \pm 0.56}(15.8\%)$ \\
& CPhet    & $3.68_{\scriptscriptstyle \pm 0.23}(4.3\%)$ \\
\cmidrule(lr){2-3}
& \cellcolor{lightblue}\textbf{\ours}
& \cellcolor{lightblue}$\mathbf{3.52_{\scriptscriptstyle \pm 0.56}}$ \\

\midrule

\multirow{5}{*}{BloodMNIST}
& SplitCP  & $4.04_{\scriptscriptstyle \pm 0.26}(23.0\%)$ \\
& FedCP-QQ & $7.76_{\scriptscriptstyle \pm 1.94}(59.9\%)$ \\
& FCP      & $4.96_{\scriptscriptstyle \pm 0.48}(37.3\%)$ \\
& CPhet    & $4.03_{\scriptscriptstyle \pm 0.35}(22.8\%)$ \\
\cmidrule(lr){2-3}
& \cellcolor{lightblue}\textbf{\ours}
& \cellcolor{lightblue}$\mathbf{3.11_{\scriptscriptstyle \pm 0.33}}$ \\

\midrule

\multirow{5}{*}{TissueMNIST}
& SplitCP  & $5.17_{\scriptscriptstyle \pm 0.23}(6.4\%)$ \\
& FedCP-QQ & $8.00_{\scriptscriptstyle \pm 1.55}(39.5\%)$ \\
& FCP      & $5.21_{\scriptscriptstyle \pm 0.23}(7.1\%)$ \\
& CPhet    & $5.03_{\scriptscriptstyle \pm 0.17}(3.8\%)$ \\
\cmidrule(lr){2-3}
& \cellcolor{lightblue}\textbf{\ours}
& \cellcolor{lightblue}$\mathbf{4.84_{\scriptscriptstyle \pm 0.40}}$ \\

\midrule

\multirow{4}{*}{RetinaMNIST}
& FedCP-QQ & $5.81_{\scriptscriptstyle \pm 3.66}(20.3\%)$ \\
& FCP      & $5.36_{\scriptscriptstyle \pm 0.51}(13.6\%)$ \\
& CPhet    & $4.85_{\scriptscriptstyle \pm 0.58}(4.5\%)$ \\
\cmidrule(lr){2-3}
& \cellcolor{lightblue}\textbf{\ours}
& \cellcolor{lightblue}$\mathbf{4.63_{\scriptscriptstyle \pm 0.57}}$ \\

\bottomrule
\end{tabular}
}
\end{table}

While marginal coverage is a prerequisite for safety, the utility of a conformal system is determined by its efficiency$-$the ability to produce the smallest possible prediction sets or intervals that still satisfy the coverage requirement. Table~\ref{tab:efficiency_comparison} reports prediction set sizes (classification) and interval lengths (regression) across all benchmarks. As a result, \ours consistently achieves the highest efficiency, producing the most compact prediction sets or intervals across all seven datasets. Improvements are particularly pronounced on heterogeneous benchmarks, demonstrating that weighted aggregation preserves the predictive strength of stronger agents without inflating uncertainty due to weaker ones.
\section{Ablation Study} \label{sec:ablation}

\begin{figure}[t]
    \centering
    \includegraphics[width=0.65\columnwidth]{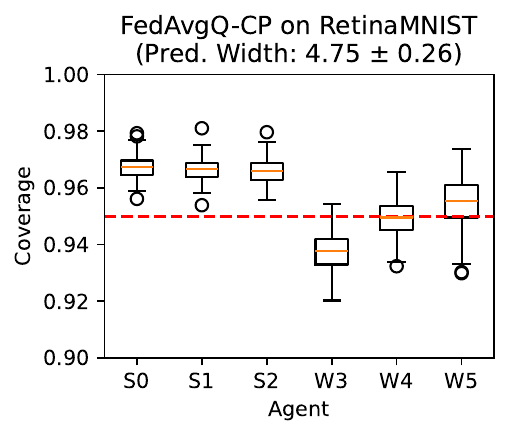}
    \caption{
    Ablation on RetinaMNIST (denoted as \textsc{FedAvgQ-CP}). Agent-level coverage for strong (\textbf{S0--S2}) and weak (\textbf{W3--W5}) agents. The red dashed line indicates the nominal 0.95 coverage level. Results represent the median $\pm$ 95\% CI over 10 independent runs (target error $\alpha = 0.05$, partition $\mathrm{Dir}(0.3)$). Unweighted quantile aggregation leads to systematic under-coverage on weak agents, underscoring the necessity of sample-size-aware aggregation in heterogeneous federated settings.
    }
    \label{fig:fedavgqcp_ablation}
\end{figure}
 
Here, we ablate the sample-size-weighted quantile aggregation in \ours by an unweighted averaging scheme, which we denote as \textsc{FedAvgQ-CP}. As shown in Figure~\ref{fig:fedavgqcp_ablation}, removing calibration-size weighting leads to systematic under-coverage on weaker agents, despite strong agents remaining near nominal levels. This confirms that sample-size-aware aggregation stabilizes the global threshold when calibration sets differ across agents.
\section{Conclusion}
In this paper, we extend CP to a realistic yet extreme FL scenario where we focus on two forms of heterogeneity, data-level (label shift or covariate shift and varying calibration set sizes) and model-level (different architectures and predictive strengths), and present \ours, a one-shot framework that aims to control coverage close to $1-\alpha$ through weighted quantile aggregation. 
We empirically validate that \ours consistently achieves near-nominal coverage across multiple diverse benchmarks,
whereas federated UQ baselines exhibit either over-coverage or severe under-coverage. Furthermore, \ours significantly improves efficiency, reducing set sizes or interval lengths compared to state-of-the-art methods. By requiring only a single communication round of two scalars per agent, \ours provides a highly scalable and privacy-preserving solution for dependable UQ in heterogeneous FL systems.

\section*{Contributions} 
     Q.-H.N conceived the idea, coded \ours, implemented analyses, and drafted the manuscript.
    J.W and W.-S.K provided feedback and proofread the manuscript. J.W and W.-S.K co-supervised the work. 
    All authors read and approved the final manuscript.

\bibliography{ref}
\bibliographystyle{plain}

\newpage

\appendix
\section*{Appendix}
\addcontentsline{toc}{section}{Appendix}

\newtheorem*{propositionrestated}{Proposition \ref{prop:dist_shift_bound} (Oracle Pooled Split Conformal Threshold)}
\section{Proof of Proposition~\ref{prop:dist_shift_bound}} 
\label{app:proof_shift_bound}

\begin{propositionrestated}
Let $P_{\mathrm{mix}}= \sum_{k=1}^M \frac{n_k}{N} P_k^{\mathrm{cal}}$ with
$N = \sum_{k=1}^M n_k$.
Let $\widehat q_{\mathrm{mix}}^\star$ denote the split conformal threshold
computed from an i.i.d. calibration sample of size $N$ drawn from
$P_{\mathrm{mix}}$ at level $\alpha$.
Then under Assumption~\ref{assump:agent_exchange},
\[
\left|
\mathbb{P}_{P_{\mathrm{test}}}
\big(
Y \in C_{\widehat q_{\mathrm{mix}}^\star}(X)
\big)
-
(1-\alpha)
\right|
\le
\sum_{k=1}^M
\frac{n_k}{N}
\, d_{TV}(P_k^{cal}, P_{\mathrm{test}})
\;+\;\frac{1}{N+1}.
\]
\end{propositionrestated}

\paragraph{Proof.}
Define the mixture calibration distribution
\[
P_{\mathrm{mix}}
=
\sum_{k=1}^M \frac{n_k}{N} P_k^{cal}.
\]

Step 1: Reference Validity under the Mixture Distribution.

Define the mixture calibration distribution
\[
P_{\mathrm{mix}}
=
\sum_{k=1}^M \frac{n_k}{N} P_k^{\mathrm{cal}}.
\]

Consider an idealized calibration procedure in which calibration scores are drawn i.i.d.\ from $P_{\mathrm{mix}}$ and the usual split conformal quantile at level $1-\alpha$ is computed from these i.i.d.\ samples.
Under the standard exchangeability assumption for i.i.d.\ samples from $P_{\mathrm{mix}}$, split conformal prediction ensures marginal validity:
\[
\mathbb{P}_{P_{\mathrm{mix}}}
\big(
Y \in C_{\widehat q_{\mathrm{mix}}^\star}(X)
\big)
\ge
1-\alpha.
\]

We use this mixture-i.i.d.\ construction as a reference model.
In the federated setting, the empirical CDF $\hat F_{\mathrm{mix}}$ is obtained from a stratified collection of samples drawn from $\{P_k^{\mathrm{cal}}\}$.
The mixture distribution serves as a population-level approximation that enables comparison between calibration and target distributions.

Step 2: Compare mixture and target distributions.

Let
\[
E = \{(X,Y): Y \in C_{\widehat q_{\mathrm{mix}}^\star}(X)\}.
\]
Then by definition of total variation distance,
\[
\left|
P_{\mathrm{test}}(E)
-
P_{\mathrm{mix}}(E)
\right|
\le
d_{TV}(P_{\mathrm{test}}, P_{\mathrm{mix}}).
\]

Step 3: Bound the mixture distance.

Because total variation is convex in its arguments,
\[
d_{TV}(P_{\mathrm{test}}, P_{\mathrm{mix}})
=
d_{TV}
\Big(
P_{\mathrm{test}},
\sum_{k=1}^M \frac{n_k}{N} P_k^{cal}
\Big)
\le
\sum_{k=1}^M
\frac{n_k}{N}
\, d_{TV}(P_{\mathrm{test}}, P_k^{cal}).
\]

Step 4: Combine.

By the triangle inequality,
\[
\left|
P_{\mathrm{test}}(E) - (1-\alpha)
\right|
\le
\left|
P_{\mathrm{test}}(E) - P_{\mathrm{mix}}(E)
\right|
+
\left|
P_{\mathrm{mix}}(E) - (1-\alpha)
\right|.
\]
By Step 2,
\[
\left|
P_{\mathrm{test}}(E) - P_{\mathrm{mix}}(E)
\right|
\le
\sum_{k=1}^M
\frac{n_k}{N}
\, d_{TV}(P_k^{cal}, P_{\mathrm{test}}).
\]
By split conformal validity under i.i.d. sampling from $P_{\mathrm{mix}}$
(using the usual non-randomized quantile choice),
\[
0 \le P_{\mathrm{mix}}(E) - (1-\alpha) \le \frac{1}{N+1},
\]
so that
\[
\left|
P_{\mathrm{mix}}(E) - (1-\alpha)
\right|
\le \frac{1}{N+1}.
\]
Combining the above bounds gives
\[
\left|
P_{\mathrm{test}}(E) - (1-\alpha)
\right|
\le
\sum_{k=1}^M
\frac{n_k}{N}
\, d_{TV}(P_k^{cal}, P_{\mathrm{test}})
\;+\;\frac{1}{N+1}.
\]

This completes the proof.

\paragraph{Oracle mixture reference.}
Proposition~\ref{prop:dist_shift_bound} analyzes an idealized
calibration procedure in which calibration samples are drawn
i.i.d. from the mixture distribution $P_{\mathrm{mix}} = \sum_{k=1}^M \frac{n_k}{N} P_k^{\mathrm{cal}}$.
This oracle construction serves as a reference model.
In the actual FL setting, calibration samples are
independent but not identically distributed across agents.
The proposition therefore provides a benchmark bound
for mixture calibration rather than a finite-sample guarantee
for the stratified federated calibration sample.

\section{Proof of Proposition~\ref{prop:agg_stability}} 
\label{app:proof_prop_2}

We first use a standard quantile perturbation argument.
Since $F_{\mathrm{mix}}$ is differentiable near $q_{\mathrm{mix}}$
and $f_{\mathrm{mix}}(v) \ge c > 0$ in a neighborhood of
$q_{\mathrm{mix}}$, the inverse function theorem implies that
for any $v$ in this neighborhood,
\[
|v - q_{\mathrm{mix}}|
\le
\frac{1}{c}
|F_{\mathrm{mix}}(v) - (1-\alpha)|.
\]

Applying this with $v = q_{\mathrm{avg}}$, we obtain
\[
|q_{\mathrm{avg}} - q_{\mathrm{mix}}|
\le
\frac{1}{c}
\big|F_{\mathrm{mix}}(q_{\mathrm{avg}}) - (1-\alpha)\big|.
\]

By definition of the mixture CDF,
\[
F_{\mathrm{mix}}(q_{\mathrm{avg}})
=
\sum_{k=1}^M \frac{n_k}{N} F_k(q_{\mathrm{avg}}).
\]

Since $F_k(q_k) = 1-\alpha$, we have
\[
F_{\mathrm{mix}}(q_{\mathrm{avg}}) - (1-\alpha)
=
\sum_{k=1}^M \frac{n_k}{N}
\big( F_k(q_{\mathrm{avg}}) - F_k(q_k) \big).
\]

Using the density upper bound $f_k \le L$,
\[
\big|F_k(q_{\mathrm{avg}}) - F_k(q_k)\big|
\le
L \,|q_{\mathrm{avg}} - q_k|.
\]

Therefore,
\[
\big|F_{\mathrm{mix}}(q_{\mathrm{avg}}) - (1-\alpha)\big|
\le
L \sum_{k=1}^M \frac{n_k}{N} |q_{\mathrm{avg}} - q_k|.
\]

Substituting $q_{\mathrm{avg}} = \sum_{j=1}^M \frac{n_j}{N} q_j$
and applying Jensen's inequality yields
\[
\sum_{k=1}^M \frac{n_k}{N} |q_{\mathrm{avg}} - q_k|
\le
\sum_{j=1}^M \sum_{k=1}^M
\frac{n_j}{N} \frac{n_k}{N}
|q_j - q_k|.
\]

Combining completes the proof.

\section{Data Statistics} \label{app:dat_stats}



We evaluate \ours on seven public benchmarks spanning
standard vision datasets and diverse medical imaging tasks.
These datasets cover heterogeneous data modalities, binary
and multi-class classification, multi-label classification,
and regression (ordinal regression) tasks, with dataset sizes
ranging from thousands to hundreds of thousands of samples.

\subsection*{Datasets Overview}

\begin{table*}[h] 
\centering
\caption{Summary of datasets used in our experiments. We include standard vision benchmarks and diverse medical imaging datasets spanning classification and regression tasks.}
\resizebox{\textwidth}{!}{
\begin{tabular}{l l l c c r r}
\toprule
\textbf{Dataset} 
& \textbf{Data Modality} 
& \textbf{Task (\# Classes/Labels)} 
& \textbf{Image Size} 
& \textbf{Channels} 
& \textbf{\# Samples} 
& \textbf{\# Train / Val / Test} \\
\midrule

MNIST 
& Handwritten Digits 
& Multi-Class (10) 
& 28$\times$28 
& 1 
& 70,000 
& 60,000 / -- / 10,000 \\

FashionMNIST 
& Apparel Images 
& Multi-Class (10) 
& 28$\times$28 
& 1 
& 70,000 
& 60,000 / -- / 10,000 \\

CIFAR-10 
& Natural Images 
& Multi-Class (10) 
& 32$\times$32 
& 3 
& 60,000 
& 50,000 / -- / 10,000 \\

DermaMNIST 
& Dermatoscope 
& Multi-Class (7) 
& 28$\times$28 
& 3 
& 10,015 
& 7,007 / 1,003 / 2,005 \\

BloodMNIST 
& Blood Cell Microscope 
& Multi-Class (8) 
& 28$\times$28 
& 3 
& 17,092 
& 11,959 / 1,712 / 3,421 \\

TissueMNIST 
& Kidney Cortex Microscope 
& Multi-Class (8) 
& 28$\times$28 
& 1 
& 236,386 
& 165,466 / 23,640 / 47,280 \\

RetinaMNIST 
& Fundus Camera 
& Ordinal Regression (5) 
& 28$\times$28 
& 3 
& 1,600 
& 1,080 / 120 / 400 \\

\bottomrule
\end{tabular}
}
\end{table*}

\subsection*{Federated Split and Heterogeneity Design}

For all datasets, we simulate a cross-silo federated learning setting with $M=6$ agents. The data splitting procedure follows three stages:

\begin{enumerate}
    \item \textbf{Global Train–Calibration Split.}  
    The official training split is further divided into a global training set (70\%) used exclusively for model training and a global calibration set (30\%) used exclusively for conformal calibration. The official test split is kept intact and used as a shared global evaluation set.
    
    \item \textbf{Dirichlet Calibration Partition.}  
    Only the global calibration set is partitioned across agents using a Dirichlet distribution with concentration parameter $\beta = 0.3$, i.e., $\mathrm{Dir}(0.3)$.  
    For classification tasks, this induces label skew across agents.  
    For regression (RetinaMNIST), we induce covariate shift by clustering features into bins and applying the Dirichlet partition over clusters.
    
    \item \textbf{Shared Training and Testing Distributions.}  
    All agents train their local predictors on the same shared global training set and are evaluated on the same shared global test set. Hence, heterogeneity is localized to the calibration stage and model architecture differences.
\end{enumerate}

\paragraph{RetinaMNIST as Regression.}

RetinaMNIST is originally defined as an ordinal classification problem with five ordered grades. In our work, we treat RetinaMNIST as a regression task by modeling the ordered labels as numeric targets and applying CQR. This formulation enables evaluation of \ours under a regression-style uncertainty quantification setting while preserving the ordinal structure of disease severity levels.

\section{Study Design} \label{app:study_design}

\paragraph{Federated Setup.}
We simulate a cross-silo federated learning system with $M=6$ agents. All experiments are repeated over $10$ independent random seeds, and we report the median and 95\% confidence interval across runs. The target miscoverage level is fixed at $\alpha = 0.05$ (nominal coverage $1 - \alpha = 0.95$).

\paragraph{Controlled Heterogeneity.}
To rigorously evaluate \ours under realistic non-IID conditions, we introduce two controlled sources of heterogeneity: data heterogeneity and model heterogeneity.

\subsection*{Data Heterogeneity (Calibration-Level Shift)}

Heterogeneity is introduced exclusively in the calibration phase. The global calibration set is partitioned across agents using a Dirichlet distribution with concentration parameter $\beta = 0.3$, i.e., $\text{Dir}(0.3)$. Smaller values of $\beta$ induce stronger heterogeneity.

\begin{itemize}
    \item \textbf{Classification Tasks.}
    For binary and multi-class classification datasets, we apply Dirichlet label skew. For each class $c$, calibration samples belonging to class $c$ are distributed across agents according to proportions drawn from a Dirichlet distribution. This produces label imbalance and heterogeneous class priors across agents.

    \item \textbf{Regression Task (RetinaMNIST).}
    For regression, we induce covariate shift instead of label skew. We cluster calibration features into $B=5$ bins using K-means clustering and apply a Dirichlet partition over these clusters. This produces heterogeneous feature distributions across agents while preserving the global test distribution.
\end{itemize}

Importantly, the training data and the global test data remain shared across agents. Only the calibration distribution differs, isolating the impact of calibration-to-test shift on conformal thresholds.

\subsection*{Model Heterogeneity (Predictive Strength and Architecture)}

To simulate realistic cross-silo deployments where institutions possess different computational resources, we designate:

\begin{itemize}
    \item Three \textbf{strong agents}: indices $\{S0, S1, S2\}$.
    \item Three \textbf{weak agents}: indices $\{W3, W4, W5\}$.
\end{itemize}

Model heterogeneity is induced through differences in training intensity and architecture:

\subsubsection*{Classification Models}

For classification tasks, agents use the following backbones:

\begin{itemize}
    \item \textbf{Strong agents: LargeCNN.}  
    A two-layer convolutional neural network with:
    \begin{itemize}
        \item Conv(32 filters) → ReLU → MaxPool
        \item Conv(64 filters) → ReLU → MaxPool
        \item Fully connected layer (128 units)
        \item Output layer with $C$ logits
    \end{itemize}
    This architecture has significantly higher representational capacity and produces more stable predictive probabilities.

    \item \textbf{Weak agents: VeryWeakLinear.}  
    A single linear layer applied to flattened image pixels:
    \[
    f(x) = W x + b.
    \]
    This model lacks convolutional inductive bias and spatial feature extraction, leading to reduced accuracy and higher predictive variance.
\end{itemize}

Strong agents are trained for $E_k = 5$ epochs, while weak agents are trained for $E_k = 1$ epoch using cross-entropy loss and the Adam optimizer.

\subsubsection*{Regression Models (RetinaMNIST)}

For the regression task, we use:

\begin{itemize}
    \item \textbf{Strong agents: LargeCNNRegressor.}  
    A convolutional backbone with:
    \begin{itemize}
        \item Two convolutional layers (32 and 64 filters)
        \item Max pooling
        \item Adaptive global average pooling
        \item Fully connected layer (128 units)
        \item Scalar output
    \end{itemize}
    This model captures spatial structure and outputs a continuous scalar prediction.

    \item \textbf{Weak agents: VeryWeakLinearRegressor\_img.}  
    A single linear layer applied to flattened image features, producing a scalar output. This architecture has minimal capacity and no convolutional structure.
\end{itemize}

Regression models are trained using mean squared error loss and the Adam optimizer. As in classification, strong agents are trained for $E_k = 5$ epochs and weak agents for $E_k = 1$ epoch.

Weak agents, due to lower capacity and reduced training, produce:

\begin{itemize}
    \item Higher predictive error,
    \item Less calibrated probability distributions,
    \item Higher-variance nonconformity scores $S(X,Y)$.
\end{itemize}

Since conformal thresholds $q_k$ are empirical quantiles of nonconformity scores, higher score variance directly increases quantile estimation variance. The asymptotic variance of a sample quantile scales on the order of $\mathcal{O}(1/n_k)$, making skewed calibration splits particularly challenging for weak agents.

This controlled imbalance creates a stringent evaluation scenario in which naive aggregation can amplify noisy thresholds. The study design therefore stresses the necessity of sample-size-aware quantile aggregation in \ours.

\end{document}